\documentclass{article}

\usepackage{microtype}
\usepackage{graphicx}
\usepackage{booktabs} 

\usepackage{amsmath}
\usepackage{amssymb}

\usepackage{hyperref}

\usepackage[accepted]{icml2020}

\usepackage{tikz}
\usetikzlibrary{shapes.geometric, arrows, backgrounds}
\usepackage{subcaption}

\icmltitlerunning{Robust Classification under Class-Dependent Domain Shift}

\begin{document}

\twocolumn[
\icmltitle{Robust Classification under Class-Dependent Domain Shift}



\icmlsetsymbol{equal}{*}

\begin{icmlauthorlist}
\icmlauthor{Tigran Galstyan}{yn2,ysu}
\icmlauthor{Hrant Khachatrian}{yn2,ysu}
\icmlauthor{Greg Ver Steeg}{isi}
\icmlauthor{Aram Galstyan}{isi}
\end{icmlauthorlist}

\icmlaffiliation{yn2}{YerevaNN}
\icmlaffiliation{ysu}{Department of Informatics and Applied Mathematics, Yerevan State University}
\icmlaffiliation{isi}{Information Sciences Institute, University of Southern California}

\icmlcorrespondingauthor{Tigran Galstyan}{tigran@yerevann.com}

\icmlkeywords{Machine learning, Robust learning, Distribution shift, Invariant representations}

\vskip 0.3in
]



\printAffiliationsAndNotice{\icmlEqualContribution} 

\begin{abstract}
Investigation of machine learning algorithms robust to changes between the training and test distributions is an active area of research. 
In this paper we explore a special type of dataset shift which we call class-dependent domain shift. It is characterized by the following features: the input data causally depends on the label, the shift in the data is fully explained by a known variable, the variable which controls the shift can depend on the label, there is no shift in the label distribution. We define a simple optimization problem with an information theoretic constraint and attempt to solve it with neural networks. Experiments on a toy dataset demonstrate the proposed method is able to learn robust classifiers which generalize well to unseen domains. 
\end{abstract}

\section{Introduction}
\label{introduction}
In many real-world applications of machine learning we experience \textit{dataset shift}, i.e. the data available in the training and inference stages come from different distributions. There has been increasing recent interest in developing machine learning models that are robust to such shifts. 

In this paper we consider a scenario when the change in the data distribution can be fully explained by some known random variable $c$. Let $p(x,c,y)$ and $q(x,c,y)$ denote the data distributions during training and testing stages. We assume they satisfy the following conditions:
\begin{align}
    p(x,c,y) &= p(x|c,y)p(c|y)p(y) \label{eq:p-factor} \\
    q(x,c,y) &= q(x|c,y)q(c|y)q(y) \label{eq:q-factor} \\
    p(y) &= q(y) \label{eq:no-label-shift} \\
    p(c|y) &\ne q(c|y) \label{eq:shift-in-c} \\
    p(x|c,y) &= q(x|c,y)  \label{eq:no-cond-shift}
\end{align}

We aim to learn a model to perform \textit{anticausal} prediction \cite{causal-anticausal}, classify $y$ given $x$, and successfully generalize to unseen relationships between $c$ and $y$. We assume we have access to $c$, but have no access to $q(x)$ at the training time. We call this setup \textit{class-dependent domain shift}.

The following artificial and real-world examples demonstrate several details of such setups.
\begin{itemize}
    \item \textbf{Colored handwritten digits.} 
    Let $y$ denote the digit, $c$ denote the color and $x$ denote the handwritten image. Consider the following generative process where we first pick the digit, then choose a color and then draw an image. During test time, we choose the digits from the same distribution, but for each digit we might choose a color from a different set. The goal is to learn a model that can predict the digit with unseen colors. Note that during testing we might encounter completely new colors, but we might also see a digit in a color that was only used for a different digit in the training phase.

    \item \textbf{Heart disease classification.} Consider a heart disease with a fixed prevalence in some population. We sample a group of patients, some of which have the disease, perform electrocardiography and extract heartbeats from the ECG data. Let $x$ denote the heartbeats, $c$ denote the patient ID, and $y$ denote the existence of the disease (a binary variable). The goal is to predict the disease from the heartbeats so that the model works for unseen patients.
\end{itemize}

Our contributions include a model designed to solve the problem defined in Eq. \eqref{eq:p-factor}-\eqref{eq:no-cond-shift}, and an experimental validation of the proposed model.

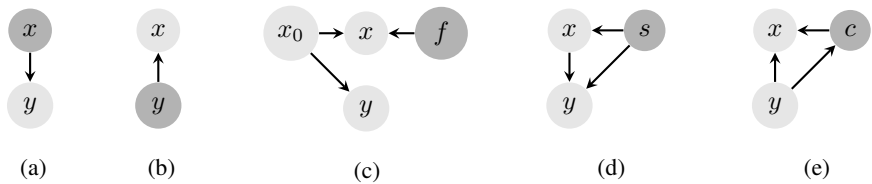
\begin{figure*}
    \tikzstyle{arrow} = [thick,->,>=stealth]

    \centering
    
    \begin{subfigure}{0.06\textwidth}
    \begin{tikzpicture}[background rectangle/.style={fill=olive!0}, show background rectangle]
        \node (x) [circle, fill=black!30] {$x$} ;
        \node (y) [circle, fill=black!10, below of=x] {$y$} ;
        \draw [arrow] (x) -- (y);
    \end{tikzpicture}
    \caption{}
    \label{fig:simple-covariate-shift}
    \end{subfigure}
    \hspace*{0.5cm}
    \begin{subfigure}{0.06\textwidth}
    \begin{tikzpicture}[background rectangle/.style={fill=olive!0}, show background rectangle]
        \node (x) [circle, fill=black!10] {$x$} ;
        \node (y) [circle, fill=black!30, below of=x] {$y$} ;
        \draw [arrow] (y) -- (x);
    \end{tikzpicture}
    \caption{}
    \label{fig:label-shift}
    \end{subfigure}
    \hspace*{0.5cm}
    \begin{subfigure}{0.18\textwidth}
    \begin{tikzpicture}[background rectangle/.style={fill=olive!0}, show background rectangle]
        \node (x) [circle, fill=black!10] {$x$} ;
        \node (y) [circle, fill=black!10, below of=x] {$y$} ;
        \node (f) [circle, fill=black!30, right of=x] {$f$} ;
        \node (x0) [circle, fill=black!10, left of=x] {$x_0$} ;
        \draw [arrow] (x0) -- (y);
        \draw [arrow] (x0) -- (x);
        \draw [arrow] (f) -- (x);
    \end{tikzpicture}
    \caption{}
    \label{fig:domain-shift}
    \end{subfigure}
    \hspace*{0.5cm}
    \begin{subfigure}{0.12\textwidth}
    \begin{tikzpicture}[background rectangle/.style={fill=olive!0}, show background rectangle]
        \node (x) [circle, fill=black!10] {$x$} ;
        \node (y) [circle, fill=black!10, below of=x] {$y$} ;
        \node (s) [circle, fill=black!30, right of=x] {$s$} ;
        \draw [arrow] (x) -- (y);
        \draw [arrow] (s) -- (x);
        \draw [arrow] (s) -- (y);
    \end{tikzpicture}
    \caption{}
    \label{fig:source-component-shift}
    \end{subfigure}
    \hspace*{0.5cm}
    \begin{subfigure}{0.12\textwidth}
    \begin{tikzpicture}[background rectangle/.style={fill=olive!0}, show background rectangle]
        \node (x) [circle, fill=black!10] {$x$} ;
        \node (y) [circle, fill=black!10, below of=x] {$y$} ;
        \node (c) [circle, fill=black!30, right of=x] {$c$} ;
        \draw [arrow] (y) -- (x);
        \draw [arrow] (c) -- (x);
        \draw [arrow] (y) -- (c);
    \end{tikzpicture}
    \caption{}
    \label{fig:ours}
    \end{subfigure}
    \caption{Several types of dataset shift visualized following the style used in \cite{dataset-shift-book}. (a) simple covariate shift, (b) prior probability shift (also known as label shift), (c) domain shift, (d) source component shift (e) class-dependent domain shift (our setup)}
    \label{fig:dataset-shifts}
\end{figure*}

\section{Related work}
\cite{dataset-shift-book} has a comprehensive analysis and a nice graphical illustration of various dataset shift scenarios. Each scenario is represented by the plot of the underlying causal graphical model, where each node corresponds to one variable, and the nodes for which the distribution might change between training and test environments are highlighted with a darker color (Fig. \ref{fig:dataset-shifts}).

Perhaps, the most widely explored type of dataset shift is called \textit{simple covariate shift}, when the joint distribution of $x$ and $y$ is factorized as $p(x,y) = p(y|x)p(x)$, $p(x)$ is different between training and test environments, while $p(y|x)$ is the same \cite{gretton2009covariate} (Fig. \ref{fig:simple-covariate-shift}). A typical example is the prediction of future events given the current state. The distribution of the states can change over time, but the way they cause future events is stable. The following examples demonstrate that the problems described above do not always satisfy the definition of a simple covariate shift. 
\begin{itemize}
    \item 
    In the Colored handwritten digits problem, additionally assume that $y \in \{4, 9\}$, $c \in \{red, blue\}$, and $p(red|4) = q(blue|9) = 1$. Now assume there is a red image $x_1$ which depicts some symbol ``between'' $4$ and $9$. If the symbol is from the training set, it is a badly written $4$, because there are no red $9$s in the training set. Similarly, it is a $9$ if it is from the test set. Hence, we have $1 = p(4|x_1) \ne q(4|x_1) = 0$, which violates the covariate shift assumption. An ideal model should not predict $4$ for $x_1$ only because $4$s happened to be red in the training set.
    \item 
    In the Heart disease classification example assume that the heartbeats $x_1$ of one sick patient from the training set are quite similar to the heartbeats of another healthy individual from the test set. We cannot assume $p(y|x_1) = q(y|x_1)$, as the heartbeat does not determine the existence of the disease. Instead, the disease affects the shape of the heartbeat, and there might be additional factors which might cause the heartbeats of healthy and sick patients to look similar.
\end{itemize}

In \textit{prior probability shift} or \textit{label shift} scenario \cite{lipton-label-shift} we assume that $p(y) \ne q(y)$, while $p(x|y)=q(x|y)$ (Fig. \ref{fig:label-shift}). Eq. \eqref{eq:no-label-shift} clearly contradicts the first assumption. If we consider the concatenation of $c$ and $y$ as an extended label, then our setup becomes similar to label shift, where only one part of the label is changed across environments. The main difference though is that we are only interested in predicting the stable part of the label without having any constraints on the distribution of the other part of the label in the test set. 

Our setup is closely related to \textit{domain shift}. \cite{dataset-shift-book} defines domain shift using a latent variable $x_0$ which corresponds to the underlying data and assumes the training algorithm has access only to the modified version of it, $x = f(x_0)$, where the modifier function $f$ changes between training and test environments. The goal is to learn a predictor which can generalize to new modifiers $f$. In many scenarios $f$ can be interpreted as a measurement of $x_0$ (e.g. photograph of an object using an RGB vs. infrared camera). One difference compared to our setup is the direction of the causality between $x$ and $y$. The other difference is that we allow the function $f$ to explicitly depend on $y$. In the example with cameras this means that the choice of the camera might depend on the object category. 


Another related scenario is called \textit{source component shift} (Fig. \ref{fig:source-component-shift}). The assumption here is that the data comes from different sources, each source has unique characteristics, and the contributions of different sources in training and test time are different. If $s$ denotes the random variable corresponding to the source, then the joint distribution is factorized as $p(x,y,s) = p(y|x,s)p(x|s)p(s)$, where the first two factors are constant among environments. 

Our setup is visualized in Fig. \ref{fig:ours}. The main difference from source component shift is that $y$ causes $x$ in our case. The other difference is the direction of the causality between $y$ and $c$ (or $s$). We believe the second difference is not critical, as in many scenarios (including in the colored handwritten digit example from the introduction) $p(c,y)$ can be factorized in both directions.

Recently, \cite{IRM} proposed a new learning algorithm called Invariant risk minimization (IRM), which can demonstrate out-of-distribution generalization for a wide range of dataset shift scenarios. It is based on the concept of distinct training environments, where the data in each environment is sampled i.i.d. from its distribution, but the causal relationships between variables can vary across environments. Our experiments with the IRM codebase did not produce good results for our setup (see Section \ref{sec:IRM}).

The methods developed for various dataset shift scenarios can be categorized into the ones which require access to the test distribution (without the labels) and ones which do not. Unsupervised domain adaptation algorithms \cite{ganin2015unsupervised} and most covariate shift adaptation methods \cite{covariate-shift-adaptation} are examples of the first category. Our setup assumes no access to the test distribution, similar to \cite{Robust-learning-HSIC}. This is also called \textit{zero-shot domain adaptation} \cite{zdda}.

The methods that attempt to handle spurious correlations between variables can be categorized into the ones which require explicit annotations of those variables and ones which discover such correlations automatically. For example, in the space of algorithms designed to learn invariant representations, the method developed in \cite{CVIB} is an example of the first category, while \cite{DSF} contains examples of the second category. We believe it is impossible to identify the spurious correlations without explicit annotations in our setup, so we assume the models do have access to the variable $c$. 

\section{The proposed method}
To obtain a classifier that will generalize to unseen $q(c|y)$ we follow the representation learning approach. We propose to learn a representation $z$ of $x$ which is rich enough for predicting $y$ but has no information about $c$ except for the information that is shared between $c$ and $y$. These ideas are formalized in the optimization problem $\max I(z:y)$ under the constraint $I(z:c|y)=0$, where $I(\cdot:\cdot)$ denotes the mutual information. This problem is relaxed to the following objective:
\begin{align}
    \max I(z:y) - \beta  I(z:c|y).
\end{align}
The first term is approximated using its variational lower bound \cite{VIB}. For the second term, we note that $I(z:c|y) = I(c:\{z,y\}) - I(c:y)$, where $I(c:y)$ is constant for the training set. Following \cite{HSIC-VAE, DSF}, we use Hilbert-Schmidt Independence Criterion \cite{HSIC} to minimize $I(c:\{z,y\})$. The resulting loss function becomes:
\begin{align}
    J  = \ & \mathbb{E} [\log p(y| z)] - \beta \text{HSIC}(c, [z, y]).
    \label{eq:shift_loss}
\end{align}

\section{Experiments on colored MNIST}
\subsection{The dataset}

Our dataset is based on MNIST images \cite{lecun-mnist}. For simplicity we take only images of digits 5, 6 and 9. We ``color'' our images the following way. First we uniformly sample and fix 12 colors. A color is defined as a $d$-dimensional vector from $[0, 1]^{d}$. We use two distinct colors per each digit for the training set and two others for the development set. Then, we add a third dimension to each image of size $d$, repeat monochrome image values across the new dimension and then multiply the image by the color \footnote{If $x_0$ is an image where white pixels are encoded by $1$, we construct colored $x$ by taking $x=c \cdot (x_0 - 0.5)$. To visualize the digit we add back $0.5$ and interpret the first three channels as RGB colors.}. We used $d=50$.

\begin{figure}
    \centering
    \includegraphics[width=0.32\textwidth]{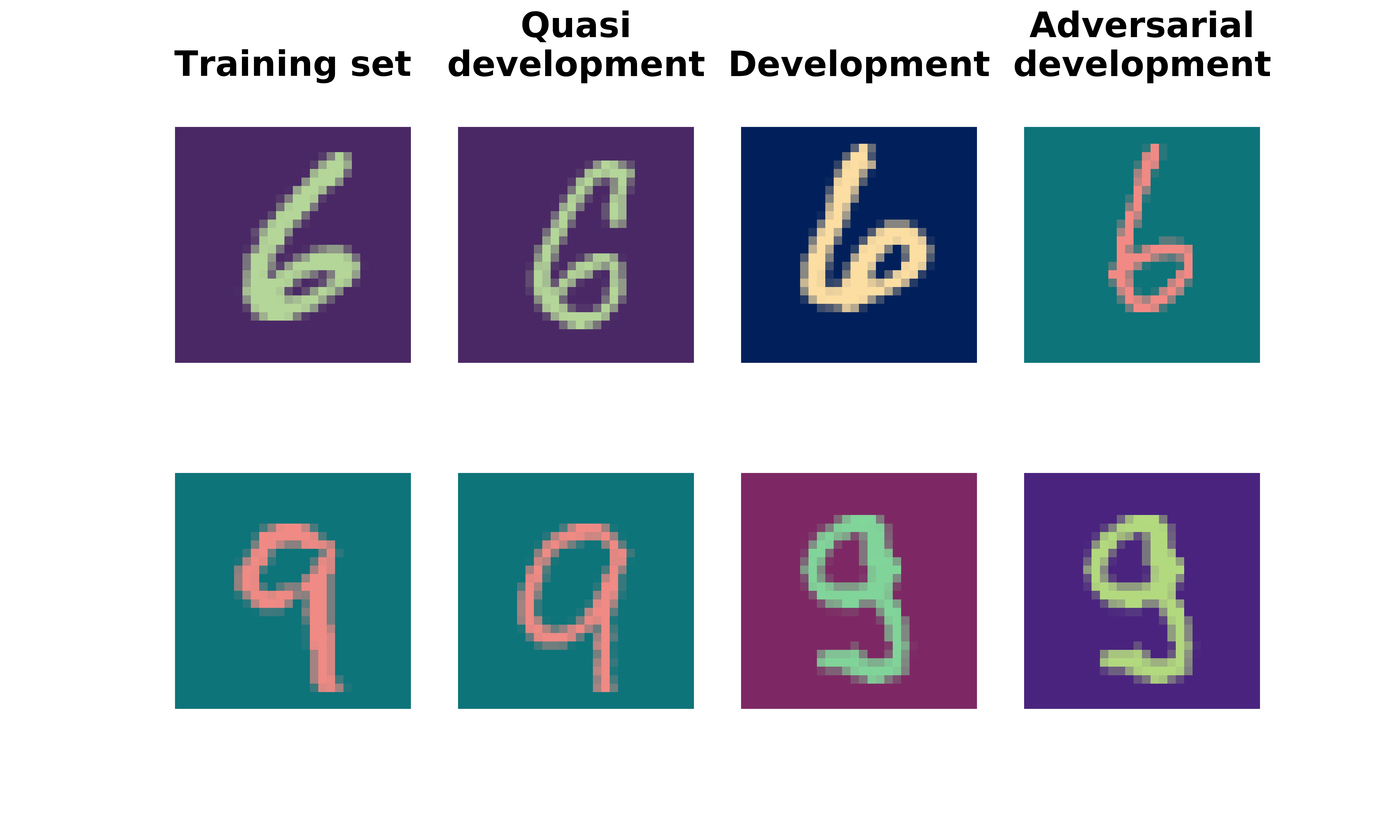}
    \caption{The training and three evaluation sets used in the colored MNIST experiment}
    \label{fig:color_mnist_sets}
\end{figure}

We evaluate the models using three datasets (Fig. \ref{fig:color_mnist_sets}). The first one is called  a \textit{quasi-development set}, where the digits have the same colors as in the training set. The second set is called \textit{development set}, where the images have completely different colors (the other 6 colors of the 12 fixed colors). And the last set is called \textit{adversarial development set}, where images have the same colors as in training set, but the colors are assigned to different digits, i.e. the color used for 6s in the training set are used for 9s etc. Any classifier that depends on the color (in contrast to the shape of the symbol) will make incorrect predictions on this set. 

\begin{figure*}
    \tikzstyle{arrow} = [thick,->,>=stealth]

    \centering
    
    \begin{subfigure}{0.24\textwidth}
    \includegraphics[width=\textwidth]{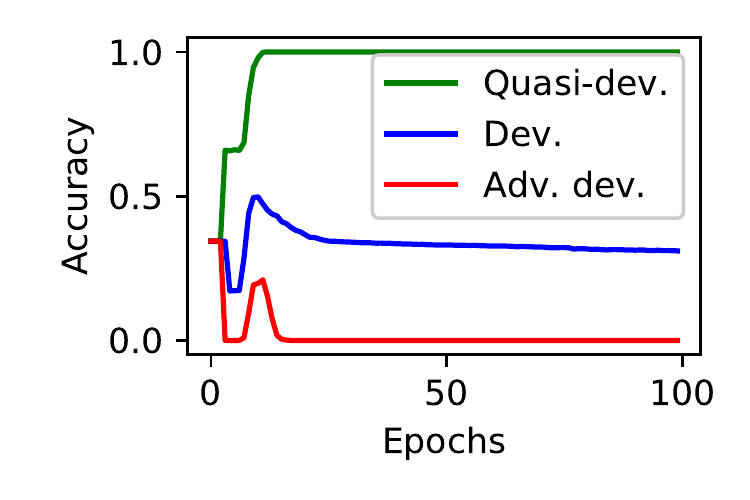}
    \caption{$\beta=0$, LR=0.0001}
    \label{fig:accuracy-base}
    \end{subfigure}
    \begin{subfigure}{0.24\textwidth}
    \includegraphics[width=\textwidth]{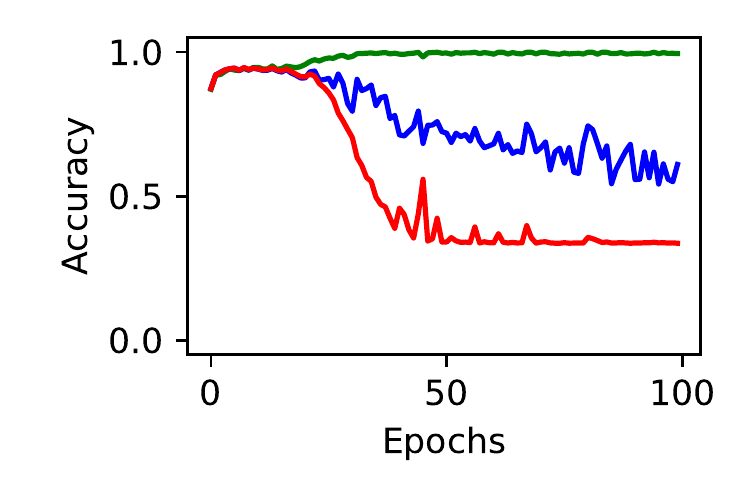}
    \caption{$\beta=2$, LR=0.001}
    \label{fig:accuracy-cond}
    \end{subfigure}
    \begin{subfigure}{0.24\textwidth}
    \includegraphics[width=\textwidth]{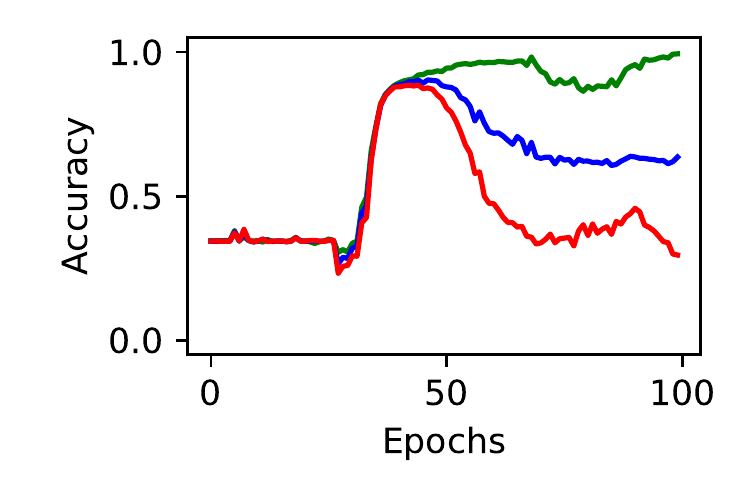}
    \caption{$\beta=2$, LR=0.001}
    \label{fig:accuracy-uncond}
    \end{subfigure}
    \begin{subfigure}{0.24\textwidth}
    \includegraphics[width=\textwidth]{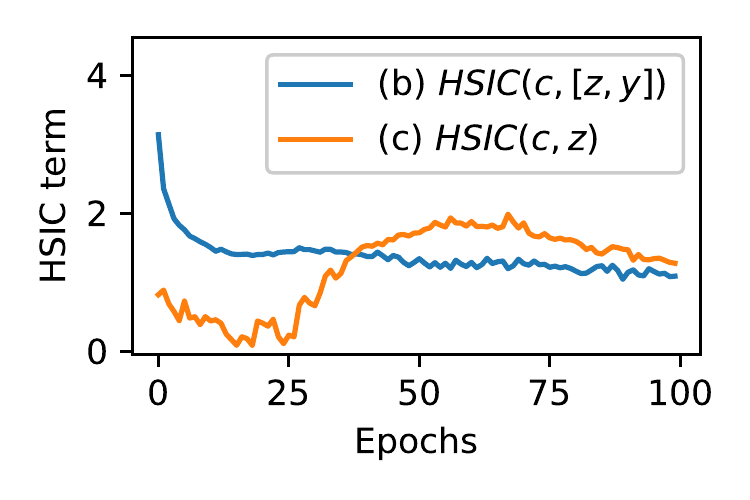}
    \caption{}
    \label{fig:hsic}
    \end{subfigure}
    \caption{Accuracies on three evaluation sets during the training process. (a) the simple baseline, (b) our method with $I(z:c|y)$, (c) the method with $I(z:c)$, (d) the values of the corresponding HSIC terms of (b) and (c).}
    \label{fig:accuracy-plots}
\end{figure*}

\subsection{Experimental setup}
To show the efficiency of our approach we compare it with a basic neural baseline. The neural network consists of a fully convolutional encoder and a linear classifier. The encoder consists of three stacked convolutional layers, and transforms $28 \times 28 \times 50$ input into a representation of size $6 \times 6 \times 5$. The only difference of our model from the basic baseline is an additional $\text{HSIC}(c, [z, y])$ term in the final loss function, where $z$ is the encoded representation of the input, $y$ is the label and $c$ is the domain-dependent variable, which is the index of the ``color'' in our case. We also perform experiments using $\text{HSIC}(c, z)$ instead of $\text{HSIC}(c, [z, y])$.

We run all experiments for fixed 100 epochs with batch size 150. We try $\beta \in \{0.1, 0.5, 1.0, 2.0, 10\}$ and use learning rates $0.01$ and $0.001$. The model selection is tricky, as we do not want to access the harder development sets in training time to approximate real-world zero-shot scenario. 

\subsection{Results and discussion}
The typical behavior of the models is shown in Fig. \ref{fig:accuracy-plots}. The simple baseline gets perfect accuracy on the quasi-development set, sometimes gets better-than-random accuracy on the development set, and always gets worse-than-random accuracy on the adversarial development set (Fig. \ref{fig:accuracy-base}). This behavior is a sign of making a prediction based on the color and not on the shape of the symbol. Confusion matrices support this explanation.

The training process of our method can be described by two distinct phases (Fig. \ref{fig:accuracy-cond}). In the first phase, the model quickly learns to classify the digits with identical accuracy on all evaluation sets (up to 95\%) and stays stable for some time. This phase is longer larger values of $\beta$ and smaller learning rates. In the second phase, the model starts to overfit on the colors used in the training set, accuracy on the quasi-development set reaches 100\%, while on the other sets it decreases. When $\beta>10$ and LR=$0.001$, the second phase does not even start in 100 epochs. 

The modified version of our method with $\text{HSIC}(c, z)$ performs similarly, except that the accuracy does not rise right from the first epoch (Fig. \ref{fig:accuracy-uncond}). It takes up to 30 epochs (faster with larger learning rates and smaller values of $\beta$) to reach 90\%+ accuracy on all evaluation sets. A possible explanation is that the term $\text{HSIC}(c, z)$ tries to minimize the mutual information $I(z:y:c)$, while the softmax term attempts to maximize the same quantity. It takes time before the softmax term wins and drives the accuracy up. On the other hand, decreasing $\text{HSIC}(c, [z, y])$ does not reduce $I(z:y:c)$. This explanation is supported by the fact that the loss term $\text{HSIC}(c, z)$ is \textit{increased} towards the end of the first phase and slowly starts to decrease in the second phase (Fig. \ref{fig:hsic}).

Unfortunately, we were not able to find a reliable way to perform model selection. If we choose the checkpoint which has the best accuracy on the quasi-development set among all hyperparameters, we get an overfitted model with 100\%, 79.77\% and	34.31\% accuracies on the three evaluation sets. If we choose the model according to the performance on the development set, then we get an optimal model with 96.01\%, 96.39\% and 95.93\% accuracies.

\subsection{Invariant Risk Minimization}\label{sec:IRM}
IRM \cite{IRM} requires the training data to be split into distinct environments. The main assumption is that the dependence of $y$ from its causal parents is the same in all environments, while the rest of the causal graph can be changed. In our setup, $y$ has no causal parent, so IRM should be applicable (at least in theory). To adapt our colored MNIST dataset for IRM, we have to define environments for the training data. We kept two labels only to bring the setup closer to the dataset used in the original IRM paper. We tested two sets of environments. 
\begin{enumerate}
    \item As we have two colors per label, we need four environments to cover our training set while having a fixed color per label in each environment.
    \item Another approach is to have two environments with all color combinations, but with different ratios of colors per label. We chose the ratio of the first and the second colors of each label to be 60:40 in the first environment and 40:60 in the second environment. This is perhaps closer to the experiments described in the original paper.  
\end{enumerate}
We performed experiments using the code provided by the authors. In all cases IRM did not perform better than the regular Empirical Risk Maximization (ERM) baseline (while the ``greyscale'' baseline worked perfectly). We additionally tried to replace the MLP used in the original code with a convolutional network, as our own method struggles to learn a generalizable model without convolutional layers, but did not see any improvement.




\section*{Acknowledgements}
We would like to thank Hrayr Harutyunyan for useful discussions. This material is based on research sponsored by Air Force Research Laboratory (AFRL) under agreement number FA8750-19-1-1000. The U.S. Government is authorized to reproduce and distribute reprints for Government purposes notwithstanding any copyright notation therein. The views and conclusions contained herein are those of the authors and should not be interpreted as necessarily representing the official policies or endorsements, either expressed or implied, of Air Force Laboratory, DARPA or the U.S. Government. T.G. and H.K. were partially supported by an EIF grant. The experiments were performed on Titan V GPUs donated to YerevaNN by NVIDIA.


\bibliography{example_paper}
\bibliographystyle{icml2020}

\appendix

\section{Other datasets}
Many real-world datasets contain samples from similar but distinct domains and sources. As most of the machine learning research is focused on the case when the samples in both the training and test sets are i.i.d., the datasets are manually shuffled and the domain-specific information is erased. As noted in \cite{IRM}, the original NIST handwritten data was collected from different writers under different conditions, but the MNIST dataset is not only shuffled, but also the information about the writers is removed. Unfortunately, this trend continues nowadays, which limits the development of robust classification methods that rely on $c$.

    The recently proposed Cells Out of Sample (COOS) dataset contains microscope images of mouse cells \cite{coos}. The cells are captured from different wells on multiple plates on different days over two years. The label is the protein which is used to highlight various parts of the cells. The paper proposes a benchmark for image classification under covariate shift. The authors prepared four test sets with varying degree of dataset shift. In fact, the way the dataset is prepared implies that the class label $y$ causes $x$, and Eq. \eqref{eq:p-factor}-\eqref{eq:no-cond-shift} describe the dataset better than the covariate shift. The original version of the dataset did not include annotations for image date, plate and well (which together can form our variable $c$). The authors of the dataset kindly agreed to add this information in a new release, and we will try our method on the dataset in future work.
    
    In \cite{2020unshuffling}, the authors attempt to ``unshuffle'' the popular visual question answering dataset VQA to obtain multiple data domains or environments. They implement a practical method inspired by Invariant Risk Minimization and demonstrate performance improvements compared to the i.i.d. baselines. We left experiments using our method on this dataset for future work.

\end{document}